\documentclass[10pt,twocolumn,letterpaper]{article}

\usepackage[pagenumbers]{cvpr} 

\definecolor{cvprblue}{rgb}{0.21,0.49,0.74}
\definecolor{present}{HTML}{C2DA98}
\definecolor{past}{HTML}{E9E899}
\definecolor{nocue}{HTML}{8BC099}
\definecolor{numericalcue}{HTML}{1F356A}
\usepackage[pagebackref,breaklinks,colorlinks,allcolors=cvprblue]{hyperref}
\usepackage{float}

\newcommand{\present}[1]{\fboxsep1pt\colorbox{present}{\textcolor{black}{\textbf{#1}}}}
\newcommand{\past}[1]{\fboxsep1pt\colorbox{past}{\textcolor{black}{\textbf{#1}}}}
\newcommand{\nocue}[1]{\fboxsep1pt\colorbox{nocue}{\textcolor{white}{\textbf{#1}}}}
\newcommand{\numericalcue}[1]{\fboxsep1pt\colorbox{numericalcue}{\textcolor{white}{\textbf{#1}}}}

\def\paperID{13} 
\def\confName{CVPR}
\def\confYear{2025}

\title{Vision language models have difficulty recognizing virtual objects}

\author{Tyler Tran, Sangeet Khemlani, J. Gregory Trafton\\
{\tt\small \{tyler.k.tran.civ, sangeet.s.khemlani.civ, greg.j.trafton.civ\}@us.navy.mil}\\
US Naval Research Laboratory\\
Washington, DC 20175 USA\\
}

\begin{document}
\maketitle

\begin{abstract}
Vision language models (VLMs) are AI systems paired with both language and vision encoders to process multimodal input. They are capable of performing complex semantic tasks such as automatic captioning, but it remains an open question about how well they comprehend the visuospatial properties of scenes depicted in the images they process. We argue that descriptions of virtual objects -- objects that are not visually represented in an image -- can help test scene comprehension in these AI systems. For example, an image that depicts a person standing under a tree can be paired with the following prompt: {\small{\texttt{imagine that a kite is stuck in the tree}}}. VLMs that comprehend the scene should update their representations and reason sensibly about the spatial relations between all three objects. We describe systematic evaluations of state-of-the-art VLMs and show that their ability to process virtual objects is inadequate.
\end{abstract}
  
\section{Introduction}
\label{sec:intro}

The ability to imagine is what permits humans to reason beyond what they perceive \cite{pearson2019human, andrews2021mapping}: they can mentally rotate images of 3D objects to imagine them in different configurations \cite{shepard1971mental}; they can animate the components of pulley systems and other physical devices \cite{hegarty1992mental, bates2015humans}; they can imagine traversals over maps, diagrams, and architectural drawings to extract relational information \cite{tversky2005visuospatial, johnson1998imagery}; they can imagine novel structures by mentally combining images of parts \cite{finke2014creative}. One theorist argues that nearly all forms of human perception engage imagination in some way \cite{brown2018infusing}; another argues that human imagination serves the central generative functions of permitting creativity, hypothetical reasoning, and counterfactual analysis \cite{abraham2016imaginative}.

New advances in AI have produced systems that appear to possess human-like imaginative abilities: for instance, vision language models (VLMs), which are systems built on pre-trained transformers architectures and coupled with vision encoders, can process imagery and text simultaneously \cite{wu2023multimodal} and some researchers have investigated how they can be used to generate imaginary scenes \cite{zhao2024imaginenav} and configurations of objects \cite{liu2024enhancing}. These models are trained on webscale image caption corpora to extract visuospatial information from out-of-distribution images, yielding possible humanlike performance on a large swath of visual tasks such as image tagging, automatic captioning, and autonomous driving \cite{hochmair2024correctness, zhang2024vision, cui2024survey}. Researchers debate the extent to which VLMs engage in robust spatial scene understanding \cite{cai2024spatialbot, fu2024scene, chen2024spatialvlm, cheng2024spatialrgpt, ghosh2024exploring, chen2024large}, especially given that they exhibit aberrant behavior that humans don't produce \cite{liu2024survey, liu2024survey2, liu2023visual}.

We argue that any general purpose scene processing system should be capable of visual imagination, i.e., imagining how an image would change given new information. Imaginative processing is particularly necessary for VLMs, which are built for ``generalist'' purposes \cite{wu2024visionllm} and vaunted for their versatility \cite{jain2024vcoder}, since they can process text and imagery concurrently. If a VLM cannot perform a variety of rudimentary imaginative tasks on input imagery, it suggests that the model cannot encode the structure of a scene in a robust and productive way.

\begin{figure}[!t]
\vskip 0.2in
\begin{center}
\centerline{\includegraphics[scale=0.23]{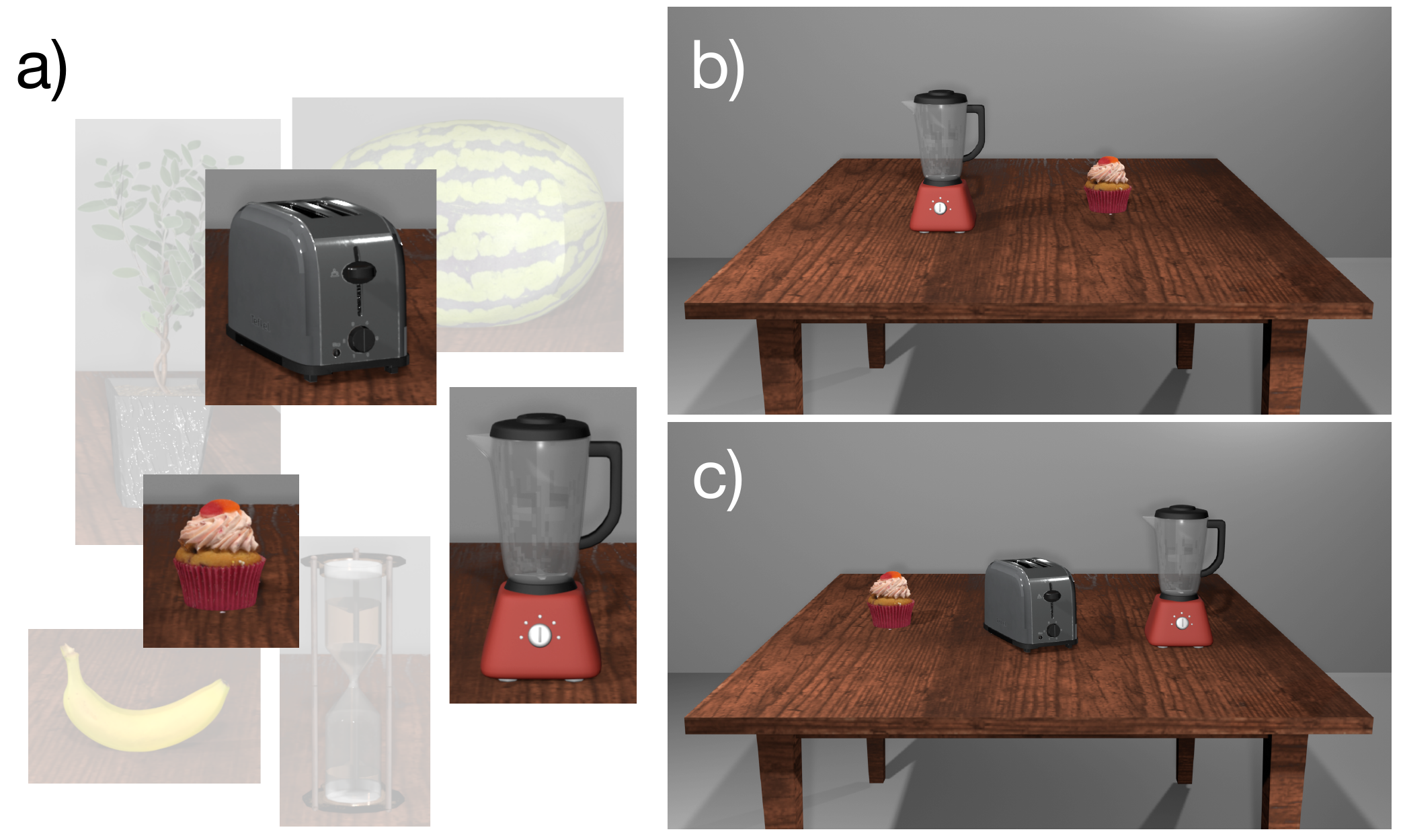}}
\caption{Examples from the \textsc{TableTest} dataset, which includes 64 individual objects (a) in various 2-object (b) and 3-object configurations (c).}
\label{figure-tabletest}
\end{center}
\vskip -0.2in
\end{figure}

\begin{table*}[h]
\begin{tabular}{p{2cm}p{7cm}p{7cm}}
\hline
\textbf{Abbreviation} & \present{Present tense}                                                 & \past{Past tense}                                                           \\ \hline
``act''                 & {\footnotesize{\texttt{\textbf{Act as though} there \present{is} a C next to the A; what items \present{are} on the table?}}} & {\footnotesize{\texttt{\textbf{Act as though} there \past{was} a C next to the A, what items \past{would be} on the table?}}} \\ 
``assume''              & {\footnotesize{\texttt{\textbf{Assume} there \present{is} a C next to the A; what items \present{are} on the table?}}} & {\footnotesize{\texttt{\textbf{Assume} there \past{was} a C next to the A; what items \past{would be} on the table?}}}\\ 
``consider''              & {\footnotesize{\texttt{\textbf{Consider that} there \present{is} a C next to the A, what items \present{are} on the table?}}} & {\footnotesize{\texttt{\textbf{Consider that} there \past{was} a C next to the A, what items \past{would be} on the table?}}} \\ 
``if''              & {\footnotesize{\texttt{What items \present{are} on the table \textbf{if} there \present{is} a C cup next to the A?}}} & {\footnotesize{\texttt{What items \past{would be} on the table \textbf{if} there \past{was} a C cup next to the A?}}}   \\ 
``imagine''              & {\footnotesize{\texttt{\textbf{Imagine} there \present{is} a C next to the A; what items \present{are} on the table?}}} & {\footnotesize{\texttt{\textbf{Imagine} there \past{was} a C next to the A, what items \past{would be} on the table?}}} \\
``pretend''              & {\footnotesize{\texttt{\textbf{Pretend} there \present{is} a C next to the A; what items \present{are} on the table?}}} & {\footnotesize{\texttt{\textbf{Pretend} there \past{was} a C next to the A; what items \past{would be} on the table?}}}\\
``suppose''              & {\footnotesize{\texttt{\textbf{Suppose} there \present{is} a C next to the A; what items \present{are} on the table?}}}  & {\footnotesize{\texttt{\textbf{Suppose} there \past{was} a C next to the A; what items \past{would be} on the table?}}} \\ \hline
\end{tabular}
\caption{Prompts used to evaluate virtual object recognition. The evaluation study varied the tense of these prompts (present vs. past) as well as whether they provided a numerical cue or not.}
\label{table-prompts}
\end{table*}

Consider a simple case of an image that depicts a person standing under a tree canopy. A VLM may be fed the following instruction: {\small{\textbf{\texttt{Imagine that a kite is stuck in the tree}}}}. In this situation, is the kite above or below the man? The kite is a \emph{virtual object}, i.e., an object within a scene that is described but not depicted. Humans have no difficulty incorporating the new information to update their mental representations of the scene, and to thereby update their understanding of the relations between the two visual objects and the one virtual object. The answer should be equally trivial for VLMs: they should respond that the kite is \emph{above} the man.

As we show, prompts concerning virtual objects can help test the multimodal capacities of VLMs and similar machine learning approaches. We describe an evaluation study of different VLMs and their capacity to recognize mentioned virtual objects in a scene. We first describe the dataset and the battery of prompts we used to benchmark virtual object recognition, and then describe the results of those evaluations -- which reveal inadequate virtual object recognition for all VLMs under investigation.

\section{Benchmarking methodology for testing virtual object recognition}

\textsc{TableTest} is a dataset of synthetic imagery for investigating relational recognition and reasoning in VLMs \cite{khemlani2025}. The dataset consists of images of 1-3 objects arranged on a table next to one another (see Figure \ref{figure-tabletest}); it uses 64 objects from the Objaverse dataset of annotated 3D objects \cite{deitke2023objaverse}, hand-scaled to ensure sensible object sizes. It contains 4,032 2-object images and ~250K 3-object images, constructed by creating all possible spatial configurations of the 64 objects.

We identified three candidate VLMs for benchmarking virtual object recognition based on the following criteria: they were recently released (post-2022), freely available, and capable of out-of-the-box, single-shot identification of the 64 objects in textsc{TableTest}. Architecture sthat matched those criteria included: Idefics2 \cite{laurenccon2024matters} (8B parameters), InstructBlip-Vicuna (7B parameters), which builds atop the BLIP architecture \cite{li2022blip}, and Llama 3.2 \cite{touvron2023llama} (11B parameters).

\begin{figure*}[!h]
\vskip 0.2in
\begin{center}
\centerline{\includegraphics[scale=1.00]{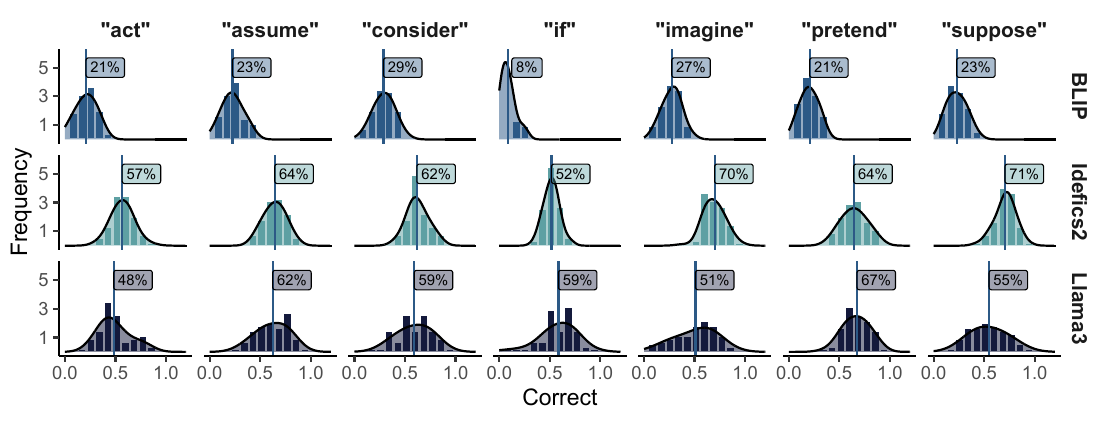}}
\caption{Proportions of accuracy from evaluations of 7 separate prompts concerning virtual object recognition in 2-object images from \textsc{TableTest}. Lines in each panel depict overall accuracies, density plots depict performance distributions across \textsc{TableTest}'s 64 objects, and bars depict histograms across those objects, as organized by whether the object served as the leftmost object in images. Humanlike performance estimated at ceiling (accuracy = 1.0).}
\label{figure-results}
\end{center}
\vskip -0.2in
\end{figure*}

The prompts used to assess virtual object recognition were minimal in nature, which allowed for systematic comparison and variation. Each prompt stipulated a hypothetical scenario that related a virtual object to one of the objects depicted in the image, e.g., {\small{\textbf{\texttt{Imagine there is a banana next to the cupcake...}}}}, and then queried for a list of the objects on the table. Successful responses are those -- and only those -- that list all three objects in any order.

Our evaluation study systematically manipulated the prompts along three dimensions: the manner in which the hypothetical was described; the tense of the prompt (past or present); and the manner in which the list of objects was queried. In theory, each of these dimensions and their variations should have no demonstrable effect on performance. We explain the study's manipulations further:

\begin{enumerate}[nosep]
  \item \textit{Prompt variations.} We subjected each VLM to 7 prompt formulations that differed in the words used to create virtual objects. For instances, prompts queried VLMs to ``assume'' or ``suppose'' that a virtual object was next to an object depicted in the image (see Table \ref{table-prompts}). The names of virtual objects were those used to describe objects in the \textsc{TableTest} dataset, and were randomized for each evaluation.
  \item \textit{Tense.} Each of the prompt variations were formulated in English using either present tense or past tense (see Table \ref{table-prompts}). Since VLMs are often trained on image caption corpora, we hypothesized incidental asymmetries in how captions are described in those corpora, and that tense could have a significant effect on performance.
  \item \textit{Numerical cues.} Half of the prompts provided on evaluations queried for a list of items in a neutral way, e.g., ``...what items are on the table?'' The other half of the prompts provided a numerical cue, e.g., ``...what three items are on the table?'' We hypothesized that numerical cues should boost performance on this task.\\
\end{enumerate}

In sum, we conducted an evaluation study in which we paired each 2-object image in \textsc{TableTest} with one of 7 different kinds of prompts $\times$ present- and past-tense versions of those prompts $\times$ queries that used numerical cues or not, yielding a total of 112,896 queries. Each of these queries were subjected to the 3 state-of-the-art VLMs. We used a fixed random seed for each evaluation and kept the temperature at 0 to ensure replicability.

\section{Evaluation study results}

Our evaluation study revealed that the VLMs under investigation produced inadequate virtual object recognition behavior for all of the 7 prompt formulations. Figure \ref{figure-results} plots the accuracies as a function of the 7 prompts and the different VLMs. In aggregate, Idefics2 produced 63\% correct responses, Llama3 produced 57\% correct responses, and BLIP produced 22\% correct responses. We calculated mean accuracies for the different objects in \textsc{TableTest} and subjected them to nonparametric analyses to assess whether these differences were statistically reliable. They revealed significant differences in performance between the three VLMs (Friedman test, $\chi^2$ = 99.08, $p < .001$). Likewise, the different prompts produced statistically reliable differences in accuracy (Friedman test, $\chi^2$ = 120.91 $p < .001$); the ``pretend'' prompt produced the best performance (51\% accuracy) and the ``if'' produced the worst (40\% accuracy). As Figure \ref{figure-results} shows, the most aberrant pattern was BLIP's performance on the ``if'' prompt, which yielded only 8\% correct responses.

\begin{figure}[!h]
\vskip 0.2in
\begin{center}
\centerline{\includegraphics[scale=1.0]{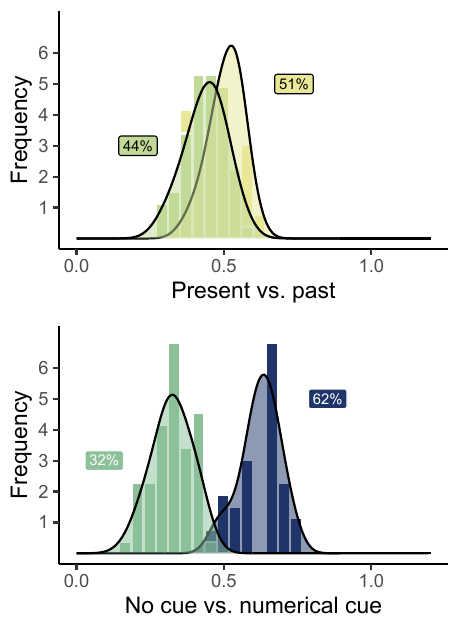}}
\caption{Proportions of accuracy from evaluations of \present{present} vs. \past{past} versions of prompts (top panel) and for \nocue{no cue} vs. \numericalcue{numerical cue} formulations; density plots depict performance distributions across 64 objects, and bars show accuracy histograms for those objects, as organized by the leftmost object in images. Humanlike performance estimated at ceiling (accuracy = 1.0).}
\label{figure-results2}
\end{center}
\vskip -0.2in
\end{figure}

The different tenses of the prompts affected performance: prompts in the past tense were more accurate than those in the present tense (51\% vs. 44\%; Wilcoxon test, $z = 6.84, p < .001$; see Figure \ref{figure-results2}, top panel). One tentative reason for this difference may be that the captions used to train VLMs -- from websites and newspapers -- may use past tense descriptions more often than present tense descriptions. Or it may be because past tense descriptions in corpora are incidentally longer and more concrete. Research into the data used to train these systems is necessary to evaluate these claims.

As hypothesized, numerical cues (``...what three items...'') reliably boosted the performance of VLMs (62\% correct with a numerical cue vs. 32\% correct without; Wilcoxon test, $z = 6.96, p < .001$). This boost may be because the cue helps the VLM consider a virtual object and its relations in a scene; or it may be for altogether trivial reasons, e.g., the prompt mentions a number and the VLM attempts to produce a response that matches any and all nouns depicted in either the scene or described in the prompt.

We underscore that all of the prompts we used have correct -- and trivial -- answers: a VLM is correct if it can describe both the depicted objects in the image and the virtual object in the prompt. And suboptimal performance should not be perturbed by irrelevant factors, such as the tense of the prompt; indeed, none of the factors we tested should have affected whether a VLM can detect virtual objects. The results therefore suggests significant limitations in the capacity for VLMs to engage in hypothetical reasoning about objects not depicted in imagery.

\section{Discussion}

We ran an evaluation study to test the imaginative capacities of three state-of-the-art vision language models (VLMs). These systems provide integrated frameworks capable of multimodal analysis by tokenizing text and images and subjecting them to transformer architectures in parallel ways. The approach has produced new capabilities, such as the robust ability to highlight and label objects in images based on natural language queries. In theory, these systems should permit simple forms of imaginative processing as well, since text embeddings can yield updated representations of embeddings of images and vice versa \cite{wu2023multimodal, zhao2024imaginenav, liu2024enhancing}. Our investigations were designed to test a rudimentary form of imagination: they tasked VLMs with imagining a new ``virtual'' object in a scene of two objects, and then queried for a list of all the objects in the scene. A system capable of updating its representations appropriately should list all three objects. Our analyses show, however, that VLMs systematically lost track of the virtual objects and were perturbed by factors that should not affect processing, such as whether prompts were in the present or past tense.

The inability to track virtual objects suggests complementary limitations on more complex tasks. VLMs could be used for many simple forms of hypothetical and imaginative reasoning by querying for the system to consider: when one object is replaced with another; when it is moved relative to another; when its size or some other property is changed, and so on. If VLMs cannot perform these tasks, they cannot be said to possess reliable visuospatial reasoning capabilities, and so their usage must be circumscribed around those tasks for which they're suited.

How could AI architectures learn to track virtual objects? Unlike contemporary AI systems, humans integrate verbal and perceptual information by building sparse, discrete, abstract ``mental models''. They construct multiple models to imagine alternative spatial configurations \cite{johnson1998imagery, knauff2013space, hegarty1992mental, tversky2005visuospatial}. Mental models discard irrelevant perceptual details \cite{cavanagh2005artist, bigelow2023non, knauff2002visual} to yield abstract, mutable structures, which permit rapid and flexible spatial reasoning of both visualizable and non-visualizable concepts \cite{ullman2018learning, cortes2021makes, kon2024spatial}. But they demand piecemeal and serial manipulation of representations \cite{khemlani2013kinematic, hegarty2010components}, which makes human reasoners slower than AI systems at many visuospatial tasks. VLMs, in contrast, leverage parallel pipelines for processing text and image embeddings holistically, but they may have difficulty integrating the resulting distributed representations in coherent ways that permit rapid updating and analysis. Systems capable of human-like imaginative processing may have to create a synthesis of these approaches, e.g., by reasoning over both distributed and discretized structures.

{
    \small
    \bibliographystyle{ieeenat_fullname}
    \bibliography{main}
}


\end{document}